\documentclass[11pt]{article} 
\usepackage{tcolorbox}
\pdfoutput=1
\usepackage{fullpage}
\usepackage{graphicx}
\usepackage{tabularx}
\usepackage{amssymb}
\usepackage{xcolor}
\usepackage[normalem]{ulem}
\usepackage{amsmath}
\usepackage{authblk}
\usepackage{url}
\usepackage{multirow}
\usepackage{booktabs}
\usepackage{adjustbox}
\usepackage{lineno}
\usepackage{soul}
\usepackage{amsmath}
\usepackage{textcomp}
\usepackage{amssymb}
\usepackage[utf8]{inputenc}
\usepackage[textwidth=1.cm, textsize=tiny]{todonotes}
\usepackage{multirow}
\usepackage{url}

\usepackage{cite}

\usepackage{amsmath}
\usepackage{array}

\def\Address{
$^{1}$U.S. Army Research Laboratory, Maryland, USA \\
$^{2}$Laboratory for Intelligent Imaging and Neural Computing, Columbia University, New York, USA \\
$^{3}$Department of Bioengineering, University of Pennsylvania, USA \\
$^{4}$Department of Physiological Brain Sciences, University of California, Santa Barbara, California, USA \\
}
\def\corrAuthor{Corresponding Author}

\def\corrEmail{nick.waytowich@gmail.com}

\begin{document}

\newpage

\title{\textbf{Compact Convolutional Neural Networks for Classification of Asynchronous Steady-state Visual Evoked Potentials}} 


\author[1,2,*]{Nicholas R. Waytowich}
\author[1]{Vernon J. Lawhern}
\author[1,3]{Javier O. Garcia}
\author[2]{Jennifer Cummings}
\author[2]{Josef Faller}
\author[2]{Paul Sajda}
\author[1,3,4]{Jean M. Vettel}

\affil[1]{U.S. Army Research Laboratory, Maryland, USA}
\affil[2]{Laboratory for Intelligent Imaging and Neural Computing, Columbia University, New York, USA}
\affil[3]{Department of Bioengineering, University of Pennsylvania, Pennsylvania, USA}
\affil[4]{Department of Physiological Brain Sciences, University of California, Santa Barbara, California, USA}
\affil[*]{Corresponding Author}

\maketitle

\begin{abstract}
\label{sec:abstract}
\textit{Objective.} Steady-State Visual Evoked Potentials (SSVEPs) are neural oscillations from the parietal and occipital regions of the brain that are evoked from flickering visual stimuli. SSVEPs are robust signals measurable in the electroencephalogram (EEG) and are commonly used in brain-computer interfaces (BCIs). However, methods for high-accuracy decoding of SSVEPs usually require hand-crafted approaches that leverage domain-specific knowledge of the stimulus signals, such as specific temporal frequencies in the visual stimuli and their relative spatial arrangement. When this knowledge is unavailable, such as when SSVEP signals are acquired asynchronously, such approaches tend to fail. \textit{Approach.} In this paper, we show how a compact convolutional neural network (Compact-CNN), which only requires raw EEG signals for automatic feature extraction, can be used to decode signals from a 12-class SSVEP dataset without the need for user-specific calibration. \textit{Main results.} The Compact-CNN demonstrates across subject mean accuracy of approximately 80\,\%, out-performing current state-of-the-art, hand-crafted approaches using canonical correlation analysis (CCA) and Combined-CCA. Furthermore, the Compact-CNN approach can reveal the underlying feature representation, revealing that the deep learner extracts additional phase- and amplitude-related features associated with the structure of the dataset. \textit{Significance.} We discuss how our Compact-CNN shows promise for BCI applications that allow users to freely gaze/attend to any stimulus at any time (e.g., asynchronous BCI) as well as provides a method for analyzing SSVEP signals in a way that might augment our understanding about the basic processing in the visual cortex.

\end{abstract}
\textbf{Keywords}: Brain-Computer Interface, EEG, Deep Learning, Convolutional Neural Network, Steady-state visual evoked potentials
\section{Introduction}
\label{sec:intro}

Evoked potentials are robust signals in the electroencephalogram (EEG) induced by sensory stimuli, and they have been used to study normal and abnormal function of the sensory cortex \cite{Regan1989a}.  The most well-studied of these are Steady-State Visual Evoked Potentials (SSVEPs), which are neural oscillations in the visual cortex that are evoked from stimuli that temporally flicker in a  narrow frequency band \cite{Herrmann2001, regan1977steady}. SSVEPs likely arise from a reorganization of spontaneous intrinsic brain oscillations in response to a stimulus \cite{bacsar1980eeg}. Paradigms leveraging SSVEP responses have been used to investigate the organization of the visual system \cite{regan1989human, silberstein2001steady}, identify biomarkers of disease and sensory function \cite{marx1986temporal, clementz2004aberrant, harding2005photic}, and probe visual perception \cite{tononi1998investigating, garcia2013near}.

The robustness of SSVEP has enabled its use as a control signal for brain computer interfaces (BCIs) that enable low-bandwith communication for individuals with catastrophic loss of motor functions, bypassing neuro-muscular pathways and establishing a communication link directly to the brain \cite{Wolpaw2002, Waytowich2016b}. In a typical SSVEP BCI, a patient/subject is presented with a grid of squares on a computer monitor, where each square contains semantic information such as a letter, number, character, or action. Superimposed on these squares are visual flicker frequencies that uniquely "tag" each square, thus mapping semantics to visual temporal frequency. As one directs their gaze and attention to a particular square with the semantic information they wish to convey, an SSVEP signal at the corresponding frequency can be measured in the EEG with dominant signals in parietal and occipital electrodes. The approach, though seemingly simple, has been important for enabling communication channels for those that are locked-in and have no other means of communication, particularly those with late stage amytrophic-lateral sclerosis (ALS) \cite{Hsu2016}. 

Central to using SSVEP, whether as a mechanism for probing vision or enabling a BCI, is the need to accurately decode and analyze the frequency information. Power spectral density analysis (PSDA), for example, is often used to identify spectral peaks in the EEG data that map to the flicker frequency of the stimulus. More recent approaches have used multivariate statistical analysis techniques, such as Canonical Correlation Analysis (CCA), that employ a template matching scheme between the EEG data and a set of hand-crafted, sinusoidal reference signals \cite{Lin2007,Bin2009, Waytowich2017a}. A recent innovation in this approach, called Combined-CCA, uses a unique combination of sinusoidal templates as well as individual template responses constructed from SSVEP calibration data to improve the standard reference signals used in the CCA \cite{Nakanishi2014a, Chen2015, Waytowich2016c}. While Combined-CCA outperforms CCA, it does so at the cost of requiring user-specific calibration data to construct the reference signals.

To enable more flexible results without the need for \textit{a priori} information, deep learning approaches were leveraged for their ability to learn robust feature representations. As such, deep learning techniques have surpassed traditional approaches that rely on manual feature extraction \cite{Schmidhuber2014, LeCun2015}. Convolutional Neural Networks (CNNs) in particular have become a very popular deep learning approach for learning rich feature representations for image classification problems \cite{Hinton2012, SimonyanZ14a, He2015, HuangLW16a}, and their ability to learn invariant features has shown promise to advance methods used in EEG signal analysis  \cite{Antoniades2016, Liang2016, ThodoroffPL16, Mirowski20091927, Langkvist2012, Schirrmeister2017b, Tabar2017, Gordon2017, Lawhern2018}. Deep learning approaches, however, typically require large amounts of training data in order to prevent over-fitting, and this requirement strongly limits their viability for SSVEP BCIs which are constrained by relatively modest sample sizes in typical BCI datasets. As a consequence, previous attempts at applying CNNs to SSVEP classification have used domain-specific representations to reduce the amount of training data required \cite{Ceocotti2011, Kwak2017, Bevilacqua2014}. These approaches utilize the fast-fourier transform (FFT) in their deep-learning models, thereby transforming EEG signals from the time-domain to the frequency-domain. These FFT-based approaches constrain the model to learn only frequency-based features which may be insufficient to capture other task-relevant information. Consequently, this approach severely hinders the value of deep learning approaches on SSVEP datasets.

In this paper, we employ a deep learning approach that allows for the discovery of the underlying representations in SSVEP signals that relate flicker to semantics and train on relatively small datasets. Specifically, our approach utilizes a recently developed deep learning model by our group that is a compact convolutional neural network (Compact-CNN) and operates on broadly-filtered EEG signals. Its compact nature allows it to operate on smaller datasets, while the convolutional structure allows for the automatic extraction of task-relevant EEG features. Our Compact-CNN was applied to a previously collected SSVEP dataset \cite{Nakanishi2015} composed of 4 s long EEG epochs of data. Without using any user-specific calibration, our Compact-CNN results in substantially better classification accuracy compared to CCA and Combined-CCA. Furthermore, the underlying feature representations constructed by our Compact-CNN revealed that the deep learner is able to extract additional phase and amplitude related features associated with the SSVEP signals. We discuss how our Compact-CNN shows promise for BCI applications that allow users to be able to freely gaze/attend to any stimulus at any time (e.g., asynchronous BCI) and augments our understanding about the underlying cortical processing in the visual cortex.

\section{Methods}
\label{sec:methods}

\begin{figure}[ht]
  \begin{center}
    \includegraphics[width=0.95\textwidth]{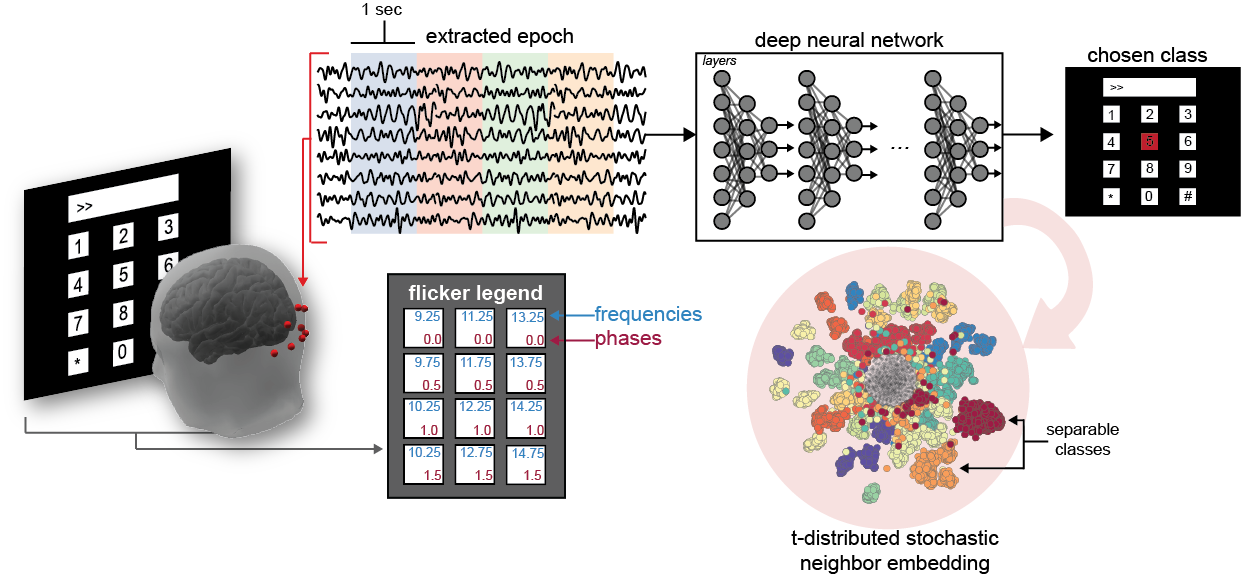}
   \caption{\emph{Methodological Overview.} Participants viewed a virtual keypad where each number flickered at a fixed frequency and phase (flicker legend). For each 4\,s trial, the participant was cued to fixate on a specific number while EEG was recorded from occipital-parietal areas (left). The trials were then divided into 1\,s segments and used as input to a deep learner, where subject-independent network models were trained and tested using a leave-one-subject-out classification procedure to identify the chosen class (the fixated number). Finally, the deep learning model activations were plotted using the t-SNE method to project the high-dimensional feature representation of the network down to two dimensions, revealing clusters that captured the diagnostic features within the test set (right).}
    \label{fig:paradigm1}
  \end{center}
\end{figure}

Ten healthy participants volunteered for an offline SSVEP BCI experiment, and their de-identified data were downloaded from a publicly available repository \cite{Nakanishi2015}. The participants sat 60\,cm away from a 27-inch LCD monitor (60\,Hz refresh-rate and 1280x800 resolution) in a dim room, and they looked at 12 flashing stimuli arranged in a 4x3 grid of 6cm x 6cm squares that represented a numeric keypad. As shown in Figure~\ref{fig:paradigm1}, the 12 SSVEP stimuli flashed at frequencies ranging from 9.25\,Hz to 14.75\,Hz in steps of 0.5\,Hz. For each individual trial, a red square was used to cue subjects to visually fixate one of the 12 stimulus squares. Each trial was 4 seconds long and was within a block of 12 trials that included each of 12 unique targets. Each subject underwent 15 blocks of these 12 trials for a total of 180 trials. During the experiment, EEG data was collected from 8 active electrodes placed over occipital-parietal areas using the BioSemi ActiveTwo EEG system (Biosemi B.V., Netherlands) with a sampling frequency of 2048\,Hz. All data were first bandpass filtered bidirectionally between 9 and 30\,Hz using a Butterworth  filter and then downsampled to 256\,Hz. The 4\,s epochs were divided in to 1\,s segments for subsequent analysis.
     
Three classification methods were compared using a leave-one-subject-out cross validation procedure where data from 9 participants were pooled together for training and the 10th participant was used for testing. In this way, user-independent models were built that did not use any training data from the test participant. The test participant was rotated for each fold so that each participant became a test participant. For classification, three methods were compared that do not require user-specific calibration: our Compact-CNN and two baseline methods derived from the conventional multivariate statistical analysis technique, Canonical Correlation Analysis (CCA). The first method tested was the calibration-free CCA that used sinusoidal reference signals \cite{Lin2007,Bin2009}, while the second was our variant of the state-of-the-art Combined-CCA method \cite{Nakanishi2014a, Chen2015} that uses transfer learning to eliminate the user-specific calibration but maintains the superior classification performance of Combined-CCA \cite{Waytowich2016c}.

Finally, the \emph{learned} representation was visualized for our Compact-CNN, addressing a scientific aim within the EEG deep learning community to understand the diagnostic features of the data \cite{Sturm2016,Schirrmeister2017b}. As depicted in Figure~\ref{fig:paradigm1}, the feature activations of the trained deep learner were shown using t-distributed stochastic neighbor embedding (t-SNE) to project individual SSVEP trials onto two dimensions \cite{Maaten2008}. The t-SNE projections were then plotted for each layer to visualize how the deep learner separates and clusters EEG trials in a projected feature-space learned by the deep network. The clusters are used to infer the diagnostic features within the test set.

\subsection{Classification Methods}
\label{sec:classifiers}

\subsubsection{EEGNet: Compact Convolutional Neural Network (Compact-CNN)}
\label{sec:eegnet}

To assess the utility of deep learning approaches for SSVEP BCI, our analysis used our Compact-CNN (the EEGNet architecture) that was designed for classifying raw EEG when only limited amounts of data are available, such as in SSVEP BCI experiments \cite{Lawhern2018}. Our Compact-CNN approach efficiently represents EEG signals in a compact manner by first performing temporal convolutions, with the convolutional kernel weights being identified from the data. The first layer of the network then performs a temporal convolution to mimic a bandpass frequency filter, a result supported by the Convolution Theorem \cite{cohen2014book}. Our approach then uses depthwise spatial convolutions that act as spatial filters to reduce the dimensionality of the data. The main benefit of depthwise convolutions is reducing the number of trainable parameters to fit, as these convolutions are not fully-connected to all previous outputs. When used in EEG-specific applications, this operation provides a direct way to learn spatial filters for each temporal filter, thus enabling the efficient extraction of frequency-specific spatial filters. The Compact-CNN also uses separable convolutions to more efficiently combine information across filters \cite{Chollet16a}. The main benefits of separable convolutions are (1) reducing the number of parameters to fit and (2) explicitly decoupling the relationship within and across outputs by first learning a kernel summarizing each output individually, then optimally merging the outputs afterwards. As described in Table 1, each convolution layer is followed by batch normalization, 2D average pooling, and dropout layers. In the first two layers, the exponential linear unit (ELU) non-linearity (as opposed to other non-linear activation functions such as sigmoids or rectified linear units) was employed since it resulted in superior performance for EEG classification \cite{Lawhern2018}. The fourth and final layer is connected to a dense layer with a softmax activation function for classification. Our Compact-CNN is trained using a categorical cross-entropy loss function (shown in Equation 1):

\begin{equation}
L_i = -\sum_{j} t_{i,j} log(p_{i,j})
\end{equation}

\noindent where $p$ are the model predictions, $t$ are the true labels, $i$ denotes the sample number, and $j$ denotes the class. The model was implemented in Tensorflow \cite{Abadi2016b}, using the Keras API \cite{chollet2015keras}. The model was trained for 500 iterations using the Adam \cite{Kingma2014} optimizer, with a minibatch size of 64 trials. The dropout probability was set to 0.5 for all layers. For this application, the model learned 96 temporal-spatial filter pairs ($F_1 = 96$, $F_2 = 96$ and $D = 1$). The complete model architecture of our Compact-CNN is shown in Table 1; our implementation can be found at https://github.com/vlawhern/arl-eegmodels.

\begin{table*}[t!]
\caption{Compact-CNN architecture, where $C = $ number of channels,  $T = $ number of time points, $F_1 = $ number of temporal filters, $F_2 = $ number of separable filters (here, $F_1 = F_2$), $D = $ number of spatial filters to learn per temporal filter and $N = $ number of classes, respectively.} 
 	\centering
 	\begin{adjustbox}{width=1\textwidth}
 	
 	\def\arraystretch{1.25}
 	\begin{tabular}{c|lllllllll}
 		\textbf{Layer} & \textbf{Layer Type} & \textbf{\# filters} & \textbf{size}    & \textbf{\# params}  &  \textbf{Output} & \textbf{Activation} & \textbf{Options} \\ \hline
 		  
         1              & Input               	&                       &                       &                                     & (C, T)                     &                     &               \\
                        & Reshape             	&                       &                       &                                     & (1, C, T)                  &                     &               \\
                        & Conv2D              	& $F_1$                 & (1, 256)              & $256 * F_1$                          & ($F_1$, C, T)              & Linear              & mode = same          \\
                        & BatchNorm           	&                       &                       & $2 * F_1$                           & ($F_1$, C, T)              &                     &               \\
                        & DepthwiseConv2D     	& D * $F_1$             & (C, 1)                & $C * D * F_1$                       & (D * $F_1$, 1, T)          & Linear              & mode = valid, depth = D, max norm = 1         \\
                        & BatchNorm           	&                       &                       & $2 * D * F_1$                       & (D * $F_1$, 1, T)          &                     &               \\
                        & Activation          	&                       &                       &                                     & (D * $F_1$, 1, T)          & ELU                 &               \\
				        & AveragePool2D       	&                       & (1, 4)                &                                     & (D * $F_1$, 1, T // 4)     &                     &               \\
                        & Dropout               &                       &                       &                                     & (D * $F_1$, 1, T // 4)     &                     &  $rate = 0.5$   \\
         2              & SeparableConv2D     	& $F_2$                 & (1, 16)               & $16 * D * F_1 + F_2 * (D * F_1)$    & ($F_2$, 1, T // 4)         & Linear              & mode = same   \\
                        & BatchNorm           	&                       &                       & $2 * F_2$                           & ($F_2$, 1, T // 4)         &                     &               \\
                        & Activation          	&                       &                       &                                     & ($F_2$, 1, T // 4)         & ELU                 &               \\
                        & AveragePool2D       	&                       & (1, 8)                &                                     & ($F_2$, 1, T // 32)        &                     &               \\
                        & Dropout               &                       &                       &                                     & ($F_2$, 1, T // 32)        &                     &  $rate = 0.5$   \\
		 3              & Flatten             	&                       &                       &                                     & ($F_2$ * (T // 32))        &                     &               \\
 		 Classifier     & Dense               	& N * ($F_2$ * T // 32) &                       &                                   & N                          & Softmax             &              
 	\end{tabular}
 	\end{adjustbox}
 	\vspace{3mm}
    \label{CNN-Model}
 \end{table*}

\subsubsection{Standard CCA}

Canonical Correlation Analysis (CCA) is a statistical analysis technique that finds underlying correlations between two multidimensional datasets. Given two multidimensional variables $X$ and $Y$, CCA seeks to find weight vectors $W_x$ and $W_y$ such that their corresponding linear projections $x = X^TW_x$ and $y = Y^TW_y$ have maximal mutual correlation. These projections are found by solving the following objective function:

\begin{align}
\max_{W_s,W_x} \rho (x,y) = \max_{W_s,W_x} \frac{E[W_x^TXY^TW_y]}{\sqrt{E[W_x^TXX^TW_x]E[W_y^TYY^TW_y]}}.
\end{align}

For SSVEP detection, CCA computes the canonical correlation between multichannel EEG data and a set of reference signals $Y_k$ composed of sines and cosines matching the fundamental and harmonic frequencies of the target stimulus. Reference signals are made for each of the $K$ frequency stimuli where each set contains $N_h$ harmonics of the fundamental frequency $f_k$. 

\begin{align}
Y_k = 
\begin{pmatrix}
\sin(2\pi f_kt)\\
\cos(2\pi f_kt)\\
\vdots\\
\sin(2\pi N_h f_kt)\\
\cos(2\pi N_h f_kt)
\end{pmatrix}
\end{align}

EEG data is canonically correlated with each set of reference signals, and SSVEP detection is made on the basis of selecting the frequency set which provides the maximum canonical coefficient, $f_k = \max_k \rho(x,f_k)$, where $f_k=f_1,f_2,... f_K$. In this analysis, $K=12$ since data was used from a 12-class SSVEP BCI \cite{Nakanishi2015} with numbers on the virtual keypad flashing at frequencies ranging from 9.25\,Hz to 14.75\,Hz in steps of 0.5\,Hz.

\subsubsection{State-of-the-art Combined-CCA}

Combined-CCA is a recent extension to the CCA method that combines the reference signals from traditional CCA with prototype responses derived from participant EEG data to better account for the imperfect match between actual brain EEG data and the approximate sine and cosine signals \cite{Nakanishi2014a, Chen2015}. In the Combined-CCA method, participants complete a calibration session before using the SSVEP BCI to collect training data for each stimulus frequency, and EEG data for each frequency is averaged across the training trials. In our recent extension of Combined-CCA \cite{Waytowich2016c}, we introduced a pooled transfer approach that averages SSVEP trial data from other subjects to eliminate a calibration session for the current user. 

To create the prototype responses, $\bar{X_k}$, data were taken from 9 of the ten subjects and averaged across trials and subjects for each stimulus frequency $k$. Using these prototype responses $\bar{X_k}$ as well as the sine and cosine reference signals $Y_k$ (Equation 3), the Combined-CCA classifies the data from the test subject, $X$, by fusing together all pairs of canonical correlations between $\bar{X_k}$, $Y_k$ and $X$ using a weighted average resulting in a combined correlation coefficient for each class, $p_k$. The frequency that maximizes this weighted correlation value is selected as the SSVEP target:

\begin{align}
f_k = \max_k \rho(k), \quad k = 1, 2, \dots K
\end{align}

\section{Results}
\label{sec:results}

Our analysis compared three classification approaches on a 12-class SSVEP BCI experiment from a publicly available dataset \cite{Nakanishi2015}. Participants were cued to fixate one of 12 numbers flickering at frequencies ranging from 9.25 to 14.75\,Hz on a virtual keypad for four seconds (Figure~\ref{fig:paradigm1}). The 4\,s epoch was divided into 1\,s segments to augment the number of trials used in a 10-fold cross-validation procedure that identified the dominant frequency in the EEG and thus determined which number was fixated.

		\begin{figure}[ht]
		\begin{center}
			\includegraphics[width=0.9\textwidth]{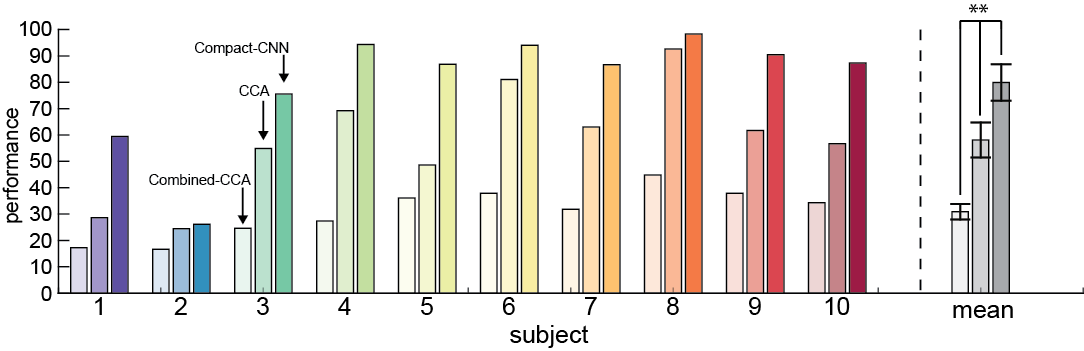}
			\caption{\emph{Classification Accuracy}. The SSVEP BCI classification accuracy is shown for each of 10 participants for three methods, Combined-CCA, CCA, and Compact-CNN. The mean accuracy across participants is shown at the right, with significant paired t-test differences between Compact-CNN and each of the multivariate methods denoted with $**$, representing a $p<.0001$. Error bars indicate SEM across participants. Nominal chance performance with 12 classes is 8.3\,\%.}
			\label{fig:fig2_perf}
		\end{center}
	\end{figure}
	
	\subsection{Compact-CNN outperforms CCA and Combined-CCA}
	
	To assess whether deep learning approaches improve classification of SSVEP signals, performance of the Compact-CNN was compared to the multivariate statistical analysis technique, CCA, and its current state-of-the-art variant, Combined-CCA. The classification accuracy for each method is shown for each of the 10 participants in Figure~\ref{fig:fig2_perf}, and results demonstrate the robust performance improvement for Compact-CNN relative to the two comparison methods. Compact-CNN proved to be particularly beneficial for subjects whose performance was notably poor when CCA based methods were used (i.e., subjects 1, 3, 9 and 10). 
	
	When evaluating the overall mean performance (Figure~\ref{fig:fig2_perf}, right), results demonstrated that Compact-CNN significantly outperformed CCA and Combined-CCA methods as indicated by paired t-tests (Compact-CNN vs Combined-CCA: $t(9) = -10.5, p<0.0001$; Compact-CNN vs CCA: $t(9) = -8.7, p<0.0001$). Surprisingly, the state-of-the-art Combined-CCA method performed significantly worse than CCA, and as subsequent analyses will show, a likely explanation is that the method is not prepared to work with asynchronous data that is not phase locked.

    \subsection{Compact-CNN extracts narrow-band frequency activity}
    
\begin{figure}[ht]
		\begin{center}
			\includegraphics[width=1\textwidth]{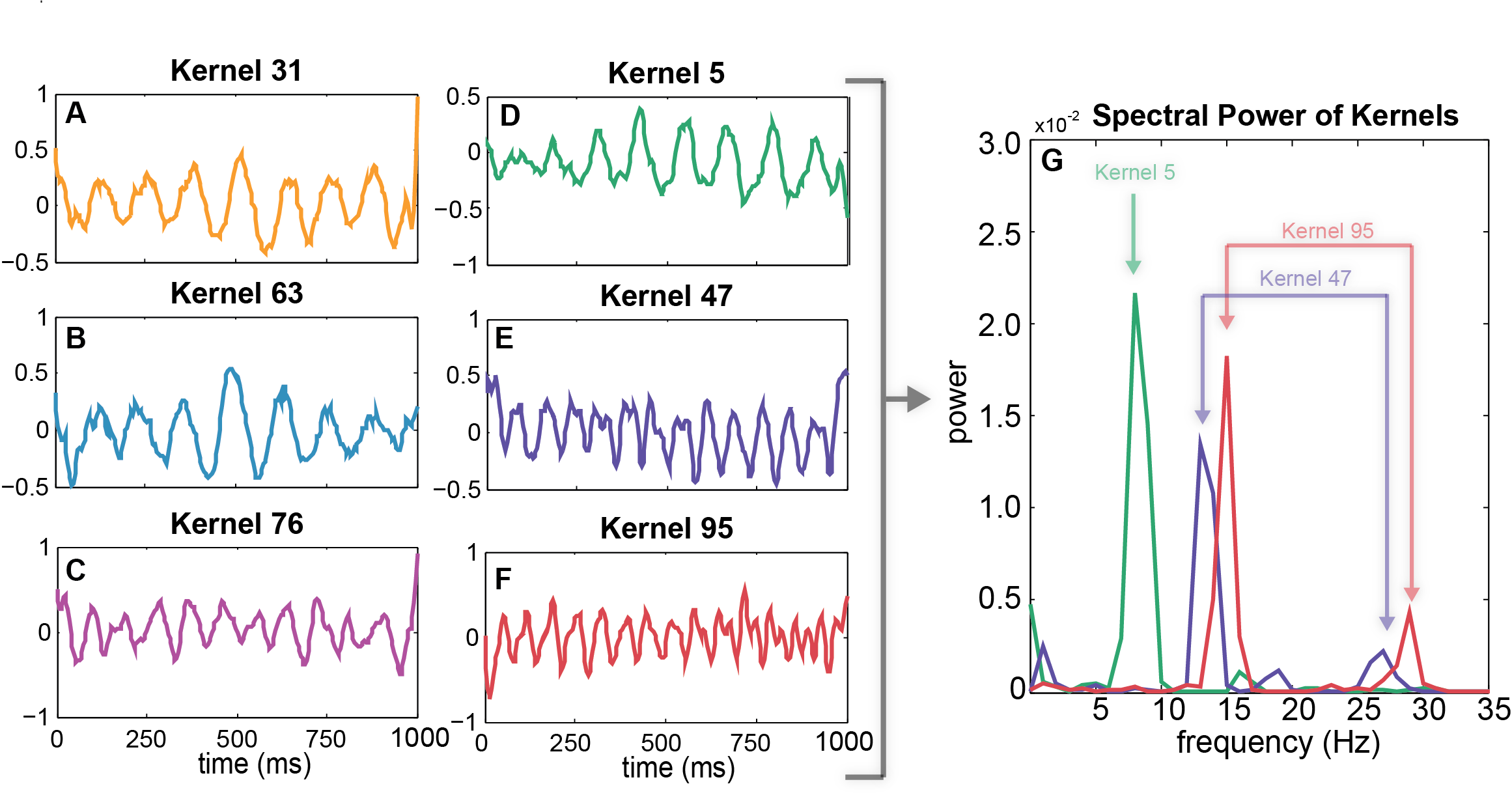}
			\caption{\emph{Visualization of Compact-CNN temporal kernels}. (A-F) Representative subset of the derived temporal kernels. In the subplots, the x-axis denotes the length (in msec) of the temporal kernel, and the y-axis denotes the amplitude of the kernel. Kernels selected based on suspected frequency separability. (G) Spectral power of the temporal kernels shown in (D-F) that depict the separability of some kernels based on the frequency content of the temporal kernels. Figure labels denote the suspected stimulus driven responses (including harmonics, e.g., kernel 95, kernel 47).}
			\label{fig:fig_kernel}
		\end{center}
	\end{figure}

    Our analysis next examined the underlying \emph{learned} representation of the Compact-CNN, investigating the diagnostic features of the data that may account for the superior performance. Our first analysis along this effort focused on interpreting the learned temporal kernel weights in Layer 1. Figure \ref{fig:fig_kernel} visualizes a subset of the temporal kernels learned by the Compact-CNN model for one randomly-selected fold and shows the corresponding spectral power of some temporal kernels. Here we see that the Compact-CNN identifies narrow-band frequency activity along a spectrum of frequencies, both slow-wave (Kernel 5, at approx. 9 Hz) and fast-wave (Kernel 95, approx. 14Hz). These frequencies closely align with that of the SSVEP experimental stimulus frequencies, suggesting that the model is capturing task-relevant oscillatory activity. 
    
	\subsection{Compact-CNN reveals differences among classes}
	
\begin{figure}[ht]
		\begin{center}
			\includegraphics[width=0.9\textwidth]{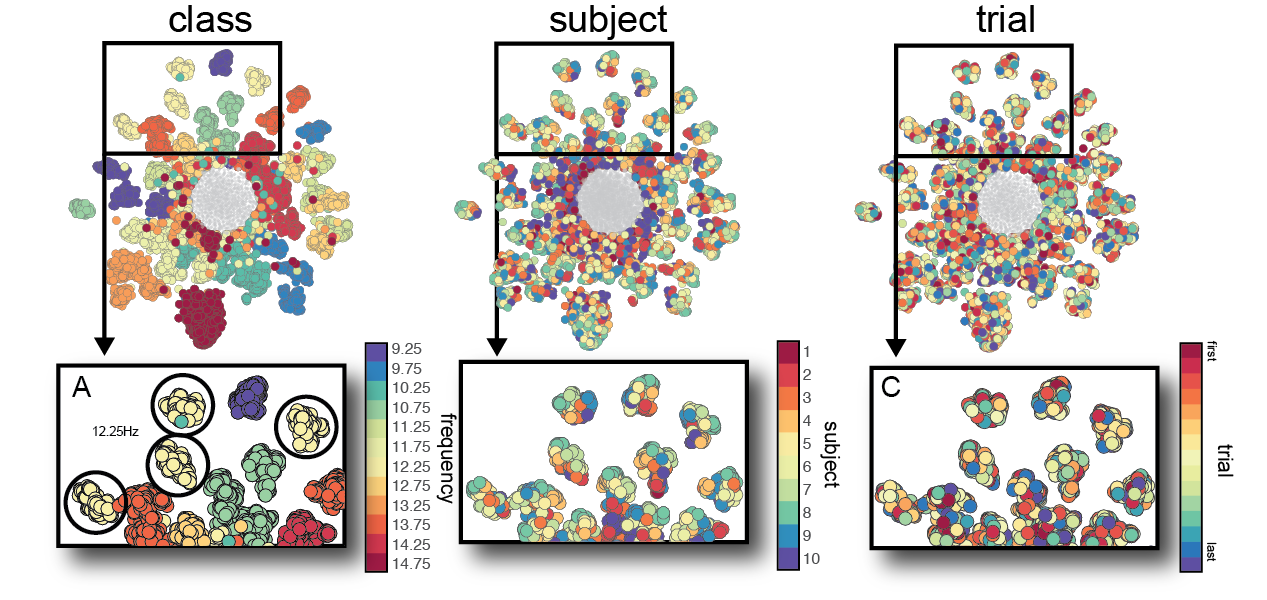}
			\caption{\emph{t-SNE of Compact-CNN Layer 3.} Each element represents a single trial and the color represents membership for three different features: (A) the stimulus class of the 12-class SSVEP numeric keypad; (B) one of the 10 participants; and (C) the trial number across the 180 trials. The bottom panels are zoomed-in sections of the top images to illustrate whether the individual clusters consist of the same colored label. Gray points indicate trials that are not visually separable and represent those least likely to be classified correctly.}
			\label{fig:fig3_tsne}
		\end{center}
	\end{figure}

	Our next analysis uses a data reduction and visualization technique (t-SNE) to investigate the hidden unit activation structure across all layers of the Compact-CNN. The activation in each layer was similar, so only the t-SNE projections of the activation in layer 3 of the Compact-CNN were plotted in Figure~\ref{fig:fig3_tsne} for all training observations (across the 9 training subjects) in fold 1.
	
	This visualization is a projection of the data down to two dimensions, allowing us to estimate discriminability among training trials and to infer the properties of the learned representation that account for classification performance. Each element of the t-SNE plot represents a single trial with a color that indicates different known features. In Figure~\ref{fig:fig3_tsne}, trials were labeled based on their class (one of 12 numbers on the virtual keypad), the participant ID (one of 10 unique individuals), and the trial order (one of 15 trial bins ordered from first to last). Only the class plot revealed coherent clusters of colored elements, indicating that the learner identified the 12 unique classes of the numeric keypad. The lack of clusters in the subject plot confirmed that the variability between individual participant brain signals does not account for the feature representation, and lack of clusters in the trial plot demonstrated that there was not a time-on-task effect (e.g., fatigue, drowsiness, inattention) that differentiated the trials across the duration of the experimental session.
	
	Next, our analysis examined the composition of the clusters themselves. Interestingly, the class plot revealed separable clusters that were colored with the same class label. In Figure~\ref{fig:fig3_tsne}A, elements colored yellow belong to the trials where the participant fixated on the number 8 that flickered at 12.25\,Hz. This reveals that the learner differentiated different trials from the same class. This unexpected separation is not predicted from the derivation of the model, as the objective function in equation (1) only seeks to separate the 12 classes. Consequently, our analysis investigated whether these within-class clusters may be related to another known feature of the dataset, namely the 1\,s segments of the original 4\,s trial epochs. 
	
	\subsection{Characterizing features of within-class clusters}
	
\begin{figure}[!ht]
		\begin{center}
			\includegraphics[width=0.8\textwidth]{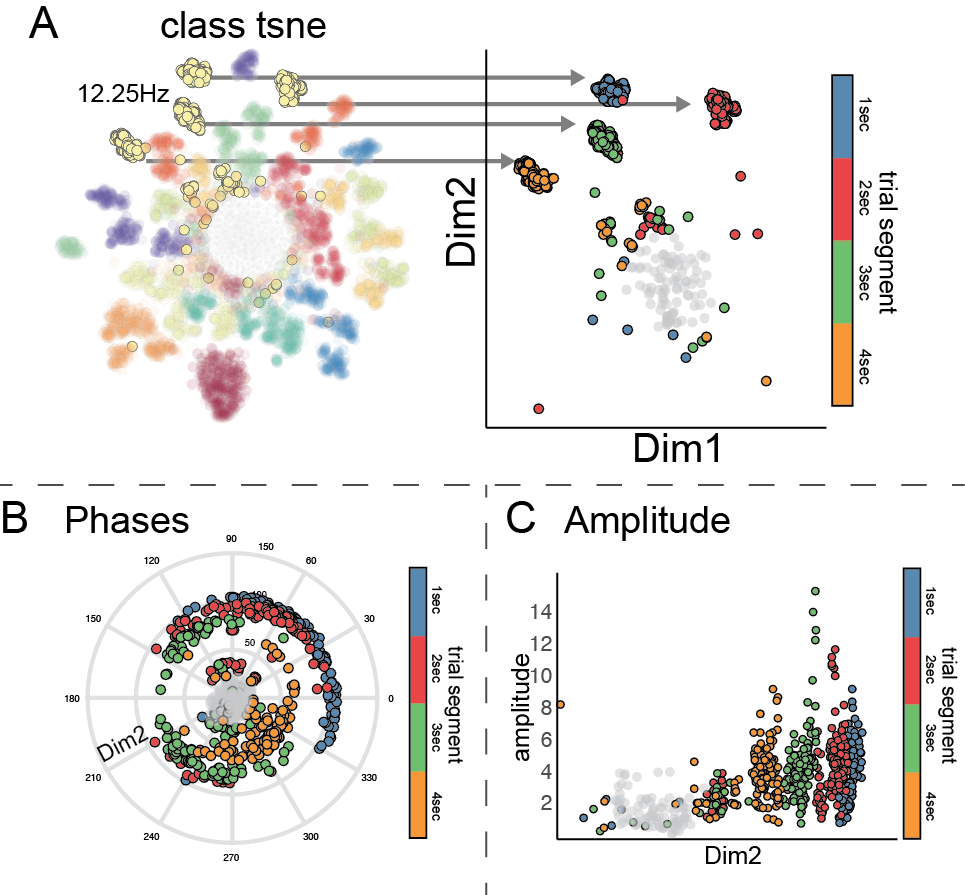}
			\caption{\emph{Phase and Amplitude of 1s Trial Segments.} (A) The yellow 12.25\,Hz clusters on the left are recolored on the right based on their respective 1\,s segment of the original 4\,s SSVEP trial, where blue is the first second, red the second, green third, and orange the fourth. (B) The polar angle plot of the estimated phase, where angle is represented in degrees within the circular plot and distance from the center of the plot represents the amplitude of Dim 2 shown in \ref{fig:fig4_phasenamp}A. (C) The estimated amplitude of the phases of \ref{fig:fig4_phasenamp}B plotted against the amplitude of Dim 2 in \ref{fig:fig4_phasenamp}A. Data shown is from channel Oz which is located over the center of the visual cortex.} 
			\label{fig:fig4_phasenamp}
		\end{center}
	\end{figure}

	In Figure~\ref{fig:fig4_phasenamp}A (right), the elements in the 12.25\,Hz class (yellow trial elements) were recolored according to their segment of the original 4\,s trial, where 0 to 1\,s is blue, 1 to 2\,s is red, 2 to 3\,s is green, and 3 to 4\,s is orange. The segments account for the separable clusters within-class. This indicates that the deep learner is influenced (and learns) EEG features other than the trained class-level differences.

	Next, our analysis characterized the signal properties of raw SSVEP responses for each trial with a phase and amplitude analysis shown in Figure~\ref{fig:fig4_phasenamp}B and \ref{fig:fig4_phasenamp}C, respectively. The radial phase plot in Figure~\ref{fig:fig4_phasenamp}B shows that the EEG trial segments have separable phases with the 3rd and 4th segments clustering at opposite phases to the 1st and 2nd segments of the trial. In contrast, Figure~\ref{fig:fig4_phasenamp}C shows similar amplitudes across the four trial segments and only Dim2 values of \ref{fig:fig4_phasenamp}A seem to denote trial segment separability.
	
\begin{figure}[!ht]
	\begin{center}
		\includegraphics[width=0.6\textwidth]{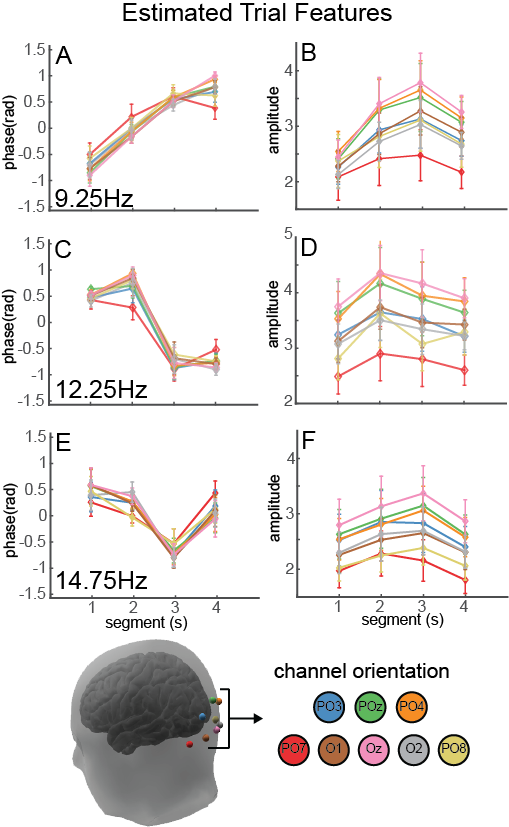}
		\caption{\emph{Mean Phase and Amplitude Across Channels.} Phase (left column) and amplitude (right column) are shown for three classes: the 9.25\,Hz class (A,B), the 12.25\,Hz class (C, D), and the 14.75\,Hz class (E, F). Electrodes are represented as a separate line and mapped onto the scalp in the legend (bottom). Error bars indicate SEM across participants.}
		\label{fig:fig5_phasenamp2}
	\end{center}
\end{figure}

	While Figure~\ref{fig:fig4_phasenamp} illustrates phase and amplitude features for the 12.25\,Hz class from channel Oz, the subplots of Figure~\ref{fig:fig5_phasenamp2} confirm that these patterns are common across other stimulus classes and channels. The average estimated phase (left column) and amplitude (right column) are depicted for the 9.25\,Hz class (top row), the 12.25\,Hz class (middle row), and the 14.75\,Hz class (bottom row). The bottom diagram in Figure~\ref{fig:fig5_phasenamp2} shows the corresponding channel layout and coloring. Phase and amplitude estimations were averaged over each trial and subject for the corresponding stimulus frequency and are shown for each electrode as colored line plots. Segment-wise differences are strongest for phase, but they can also be seen for amplitude features across all EEG channels. Here, only a few classes were illustrated for brevity, but this pattern is consistent across classes and scalp locations.

\section{Discussion}
\label{sec:discussion}

In this research, our analysis investigated whether deep learning approaches could improve performance in classification of SSVEP BCI trials over state-of-the-art methods. While CNNs have shown promise for feature extraction and classification of EEG signals across a number of domains \cite{Antoniades2016, Liang2016, ThodoroffPL16, Mirowski20091927, Langkvist2012, Schirrmeister2017b, Tabar2017, Gordon2017}, there has been very little work in applying deep learning to classification of SSVEP signals \cite{Kwak2017, Ceocotti2011, Bevilacqua2014}. We demonstrate on a representative data set that our Compact-CNN approach can outperform both CCA and Combined-CCA. Our approach improved classification performance on a group level, but also in individuals who had particularly poor performance when conventional methods were used. An inspection of the \emph{learned} representation used for classification in layer 3 of the Compact-CNN revealed activation clusters that differentiated the 12 classes of numeric stimuli in the virtual keypad. In contrast, the clusters did not differentiate individual participants from one another or the trial order that may arise from time-on-task fatigue, inattention, or other nuisance variables. The specificity of the representation in layer 3 indicates that the superior performance was not dependent on task-irrelevant features (participant, time-on-task), indicating that the Compact-CNN did not merely overfit the data in the training phase. Instead, the superior performance of the Compact-CNN likely arose from diagnostic trial features that the deep learner identified were robust in the data, namely within-class separability.

Unexpectedly, the activation in layer 3 revealed clusters of trials from within the same class, even though the 12-class classification task did not require any within-class distinction. After further investigation of the trials in the clusters, results revealed that the variability in phase across the 1\,s segments of the original 4\,s SSVEP trial accounted for the within-class cluster differentiation, and the phase differences across the four 1\,s segments were robust across channels and almost all of the stimulus classes. Although unexpected, these within-class clusters highlight the strength of the deep learning approaches to learn diagnostic features directly from the data. 

Our original motivation for the 1\,s segments arose from the aim to provide more training trials for the Compact-CNN. However, the segments led to two unexpected, but likely related, results: (1) the identification of phase variability across the EEG response in the original 4\,s SSVEP trial epoch and (2) the substantial performance impairment in classification accuracy for the current state-of-the-art method, Combined-CCA. In general, CCA approaches leverage domain knowledge to extract relevant SSVEP features embedded in the EEG responses, and our results suggest a limitation that may arise from assuming a particular stimulus frequency and phase. Most multivariate approaches, including Combined-CCA, have thus far only proven effective under system-paced or synchronized SSVEP paradigms where precise stimulus onset information is known ahead of time. Our results suggest that this approach may constrain the success of these methods to synchronous SSVEP paradigms, while our Compact-CNN approach shows promise for asynchronous SSVEP classification where the users would be able to freely gaze/attend to any stimulus at any time. Here, we discuss how our Compact-CNN approach demonstrates the value of deep learning for SSVEP BCI, provides a potential innovation for asynchronous BCIs, and augments our understanding about the underlying cortical processing in the visual cortex.

\subsection{Deep learning allows characterization of EEG features}

Deep learning approaches allow computational models to learn representations of data with multiple layers of abstraction. This in turn can allow for the discovery of rich features or structure within large datasets. However, since these layers of features are learned instead of designed by human engineers, interpreting the meaning of these features remains a critical challenge in the field of deep learning \cite{Goodfellow2014, Nguyen2014}. In this paper, our analysis utilized the representational learning capabilities of multilayer CNNs and show that the model learned interpretable features from EEG data that are consistent with the SSVEP literature. The Compact-CNN identifies phase features that are invariant to both subject and trial (time-on-task) differences. 

Our results extend previous studies that have applied deep learning to SSVEP BCI paradigms. These prior approaches leveraged domain specific knowledge when constructing their feature extraction layers \cite{Kwak2017, Ceocotti2011,Bevilacqua2014}. The first approach applied a CNN for the classification of SSVEP signals by using the fast-fourier transform (FFT) to convert time-domain representations into frequency-domain representations \cite{Ceocotti2011}. The second, more recent approach transformed SSVEP data into the frequency domain as a pre-processing step before classification with a CNN \cite{Kwak2017}. A similar approach was taken in \cite{Bevilacqua2014}, where the authors used an FFT-based hidden layer for classifying SSVEP signals. These approaches incorporate domain-specific knowledge of the stimulus frequencies by utilizing the FFT to transform raw time-domain EEG signals to the frequency domain since spectrally focal power differences are known to be discriminative features in the context of SSVEP-based decoding. All these approaches incorporate frequency specific knowledge, but no phase information. In contrast, our Compact-CNN operates on broadly-filtered EEG signals, and the relevant features (frequency and phase) are learned and extracted directly from these input signals. This end-to-end approach might be particularly advantageous for use-cases where domain-specific knowledge is unavailable.

Additionally, our results demonstrated that the Compact-CNN was able to discover deeper structure within the SSVEP dataset in the form of phase information that reflects features of the 1\,s segments of the original 4\,s SSVEP trial epochs. This learned structure is present in the t-SNE projections of the model activations in the hidden layers. Thus, the Compact-CNN provides a powerful tool for interpreting intrinsic structures within EEG datasets, and critically, this type of structure discovery would not be possible using standard linear methods, such as CCA or Combined-CCA. 

\subsection{Potential benefits of Compact-CNN-based asynchronous classification}

SSVEP signals can be generated and classified using either synchronous or asynchronous temporal coding paradigms~\cite{Mason2007}. In synchronous paradigms, the onset and offset of the user gaze towards the stimulus are known: the user is cued to look at a flashing stimulus in a specified time-window so that the user's gaze is synchronized with the onset of stimulus flashing. In contrast, asynchronous SSVEP paradigms allow users to freely gaze/attend to any stimulus at any time. Asynchronous paradigms are generally more flexible and are more amenable to practical BCI. For some applications, asynchronous temporal coding paradigms can make BCI interaction faster, more intuitive, and/or ergonomical as the user can directly communicate a specific command of choice instead of waiting for the respective system cues in synchronized approaches. EEG-based decoding however, is typically more difficult in asynchronous SSVEP paradigms, particularly if phase information of the stimulus is not known.

Here, our division of the 4\,s SSVEP trial epoch into 1\,s segments simulates an asynchronous SSVEP BCI from a discretized stimulation paradigm. These segments approximate a less-controlled paradigm wherein a participant may randomly attend to a stimulus frequency. A segment length of 1\,s represents a fast-paced BCI design where classifications can be made every second. On this data set, the Compact-CNN method outperformed state-of-the-art approaches in simulated asynchronous operation. The Compact-CNN learner extracted relevant frequency and phase features directly from the broadly-filtered EEG signals even when specific phase information of the SSVEP trials was unknown. This improvement could potentially benefit a variety of asynchronous SSVEP BCI applications including control of wheelchairs~\cite{Diez2013, Waytowich2017a}, neuroprosthetics~\cite{Mueller-Putz2006}, exoskeletons~\cite{Kwak2015}, or interaction with virtual~\cite{Legeny2011, Waytowich2017b} or augmented reality~\cite{Faller2010}. The observed performance improvement may also make this asynchronous approach attractive for applications like spellers~\cite{Chen2015}, where synchronous paradigms have been traditionally preferred for their higher information bandwidth at the cost of lower flexibility.  

Additionally, achieving calibrationless BCI classification is paramount for developing more practical BCI systems as it eliminates the need for time-consuming training sessions for the user \cite{STIG}. Using a leave-one-subject out transfer protocol, our Compact-CNN approach outperformed the baseline CCA approaches without any user-specific calibration. This is further reflected in t-SNE projections which indicate that our deep learning model is able to learn subject invariant features, which is critical for transfer across subjects. Based on its promise for asynchronous BCI paradigms and the fact that user-specific calibration is not required, this Compact-CNN approach gives BCI designers additional options to tailor SSVEP BCI systems to the needs of specific user-groups or other requirements.

\subsection{Detecting frequency and phase information with the Compact-CNN}
The Convolution Theorem states that convolutions of signals in the time-domain relate to multiplication in the frequency-domain. Since the first layer of our Compact-CNN is a temporal convolution (whose weights need to be learned from the data), our network has the capability to learn frequency-specific temporal filters, including EEG features as shown in our previous work \cite{Lawhern2018}. We showed in Figure \ref{fig:fig_kernel} that the Compact-CNN model is capable of extracting narrow-band task-specific slow-wave and fast-wave frequency activity. We believe that our network is also capturing information correlated to frequency through the use of the average-pooling layer in Layer 1 of our model. The sequence of operations in Layer 1 (temporal convolution, ELU non-linearity then average-pooling) is similar to the methodology of \cite{Pfurtscheller1999} for calculating event-related synchronization and desynchronization features. In their work, they narrow-band filter, then square, then average over a moving window the signal to obtain an estimate of frequency power. The similarity of this approach to the operations in Layer 1 of the Compact-CNN suggests that our model is calculating features at least correlated to that of frequency power.

In addition, convolutional neural networks have the property of equivariance to local translation \cite{Goodfellow2016, Wiatowski2018, Mallat2016}. In particular, when the data is a time-series signal, local translations correspond to phase offsets in the signal. This means that the output of the convolution operation (a dot product between the convolutional kernel and the signal) will preserve the phase information in the signal relative to the phase of the convolutional kernel (i.e., when the convolutional kernel is in phase or in anti-phase with the data). The ability of convolutional neural networks to capture translation-invariant features (in the case of a time-series signal, phase-invariant) was proven in  \cite{Wiatowski2018}; it was also shown that pooling layers and model depth were critical to achieve this (additional theoretical treatment of convolutional neural networks and their ability to extract translation-invariant features can be found in \cite{Mallat2016}). It is through this property that our Compact-CCN approach was able to extract phase information from the signals of interest. Figure \ref{fig:fig4_phasenamp}, which shows a t-SNE projection of the hidden unit activations in layer 3, provides additional evidence that our network can capture this phenomena.  

\subsection{Phase discrimination for vision neuroscience}

While this paper largely focuses on the relevance for BCI applications, the Compact-CNN's ability to detect phase holds incredible potential to augment our understanding of visual processing in common SSVEP experimental paradigms. For example, SSVEP experimentation has revealed the temporal dynamics and spatial constraints of spatial attention when divided across locations, refining theories of how we attend to specific locations in our environment \cite{itthipuripat2013temporal}. Likewise, phase information about oscillations of alpha activity within coordinated populations of neurons in the visual cortex has been shown to predict whether stimuli will be detected in the environment, suggesting that phase information drives coordinated excitation that facilitates perception or synchronized inhibition that prevents stimulus detection \cite{hanslmayr2007prestimulus, mathewson2009see, Brooks2016}.

Here, the Compact-CNN identified phase information in the SSVEP response that was irrelevant for the 12-class classification but robust in the neural response, both across stimulus frequencies as well as channels. The ability to detect nuanced phase variability in EEG signals holds incredible potential to augment our understanding of the role of phase information in neural processing more generally. Variability in phase measured with EEG has been shown to modulate the neural response to a stimulus \cite{barry2004event, makeig2002dynamic, jansen1991effect, haig1998prestimulus}, influence reaction times \cite{callaway1960relationship, dustman1965phase}, and determine whether multisensory input is bound as a coherent percept \cite{van2014multisensory}. Phase information within the neural response has been suggested to carry precise informational content \cite{hopfield1995pattern, jensen2000position} and may even be a component of the \emph{neural code} \cite{perkel1968neural}. Our results demonstrate that a deep learner may provide an innovative way to reveal robust phase variability between stimuli that underlie the intricate coordination of neural activity that gives rise to cognition.

\subsection{Limitations}

Within this dataset, the impaired performance of Combined-CCA likely arose from insufficient representation of the precise phase information across the 1s segments. Future research could investigate whether Combined-CCA could be adapted or used differently to better handle asynchronous operation. For example, provided that the BCI has knowledge of the phase of the stimulus, the template could be shifted so that its phase matches the phase of the stimulus. This may improve performance of Combined-CCA in an asynchronous setup. It is noteworthy though, that phase information of the stimulus is not available in all applications. Our Compact-CNN does not require phase information, suggesting its versatility for both experimental use and novel BCI applications. 

Finally, our experimental setup did not include a non-control state, which is typically utilized in self-paced BCI systems. Future work could investigate the efficacy of our method in the presence of a non-control state in a closed-loop setting, which could lead to even more flexible and intuitive SSVEP-based human machine interaction.

\section*{Acknowledgments}
 This research was supported by mission funding to the U.S. Army Research Laboratory and through Cooperative Agreement Number W911NF-10-2-0022. Support was also provided by the NSF under grant IIS-1527747. The views and conclusions contained in this document are those of the authors and should not be interpreted as representing the official policies, either expressed or implied, of the U.S. Government. The U.S. Government is authorized to reproduce and distribute reprints for Government purposes notwithstanding any copyright notation herein.

\bibliographystyle{IEEEtran}
\bibliography{ssvep_smc.bib}

\end{document}